# Mouse Simulation Using Two Coloured Tapes

Kamran Niyazi[1], Vikram Kumar[2], Swapnil Mahe[3] and Swapnil Vyawahare[4]

Department of Computer Engineering, AISSMS COE, University of Pune, India

kamran.niyazi@gmail.com[1]
vikrambag.kumar@gmail.com[2]
swapnilmahe@gmail.com[3]
swapnilvyawahare@gmail.com[4]

### ABSTRACT

*In this paper, we present a novel approach for Human Computer Interaction (HCI) where, we control cursor movement using a real-time camera. Current methods involve changing mouse parts such as adding more buttons or changing the position of the tracking ball. Instead, our method is to use a camera and computer vision technology, such as image segmentation and gesture recognition, to control mouse tasks (left and right clicking, double-clicking, and scrolling) and we show how it can perform everything as current mouse devices can.*

*The software will be developed in JAVA language. Recognition and pose estimation in this system are user independent and robust as we will be using colour tapes on our finger to perform actions. The software can be used as an intuitive input interface to applications that require multi-dimensional control e.g. computer games etc.*

## 1. INTRODUCTION

One of the important challenges in Human Computer Interactions is to develop more intuitive and more natural interfaces. Computing environments presently are strongly tied to the availability of a high resolution pointing device with a single, discrete two dimensional cursor. Modern Graphical user interface (GUI), which is a current standard interface on personal computers (PCs), is well-defined, and it provides an efficient interface for a user to use various applications on a computer. GUIs (graphical user interfaces) combined with devices such as mice and track pads are extremely effective at reducing the richness and variety of human communication down to a single point.

While the utility of such devices in today's interfaces cannot be denied, there are many users who find that the capability of GUI is rather limited when they try to do some tasks by using gestures. There are opportunities to apply other kinds of sensors and techniques to enrich the user experience of such users. For example, video cameras and computer vision techniques may be used to capture many details of human shape and movement. The shape of the hand may be analyzed over time to manipulate an onscreen object in a way analogous to the hand's manipulation of paper on a desk. Such an approach may lead to a faster, more natural, and more fluid style of interaction for certain tasks.

Ubiquitous computing is devoted to changing the relationship between humans and the computers with which we interact, towards allowing computers to become invisible and recede into the periphery of people's lives.

Our project, Mouse Simulation Using Two Coloured Tapes is an attempt in ubiquitous computing. Here, we will be using two colour tapes on our fingers. One of the tapes will be used for controlling cursor movement while the relative distance between the two coloured tapes will be used for click events of the mouse. Thus, the system will provide a new experience for users in interacting with the computer.

## 2. RELATED WORK

A lot of research is being done in the fields of Human Computer Interaction (HCI) and Robotics. Researchers have tried to control mouse movement using video devices for HCI. However, all of them used different methods to make mouse cursor movement and clicking events.

One approach, by Hojoon Park [1] used index finger for cursor movement and angle between index finger and thumb for clicking events. Also, Erdem et al [2], used finger tip tracking to control the motion of the mouse. A click of the mouse button was implemented by defining a screen such that a click occurred when a user's hand passed over the region. Another approach was developed by Chu-Feng Lien [3]. He used only the finger-tips to control the mouse cursor and click. His clicking method was based on image density, and required the user to hold the mouse cursor on the desired spot for a short period of time. Paul et al [5], used another method to click. They used the motion of the thumb (from a 'thumbs-up' position to a fist) to mark a clicking event thumb. Movement of the hand while making a special hand sign moved the mouse pointer. S Malik [4] developed a real-time system that can track the 3D position and 2D orientation of the thumb and index finger of each hand without the use of special markers or gloves. His system could be used for single pointing and pinching gestures. In robotics Asanterabi Malima et al. [6] developed a finger counting system to control behaviour of a robot.

A study of the existing systems for on-screen choice selection reveals that people are still limited to the use of devices such as mouse, touchpad, joystick, trackball and touch screen. All these devices need contact of hand with them. Our proposed approach is touch-free.

## 3. SYSTEM OVERVIEW

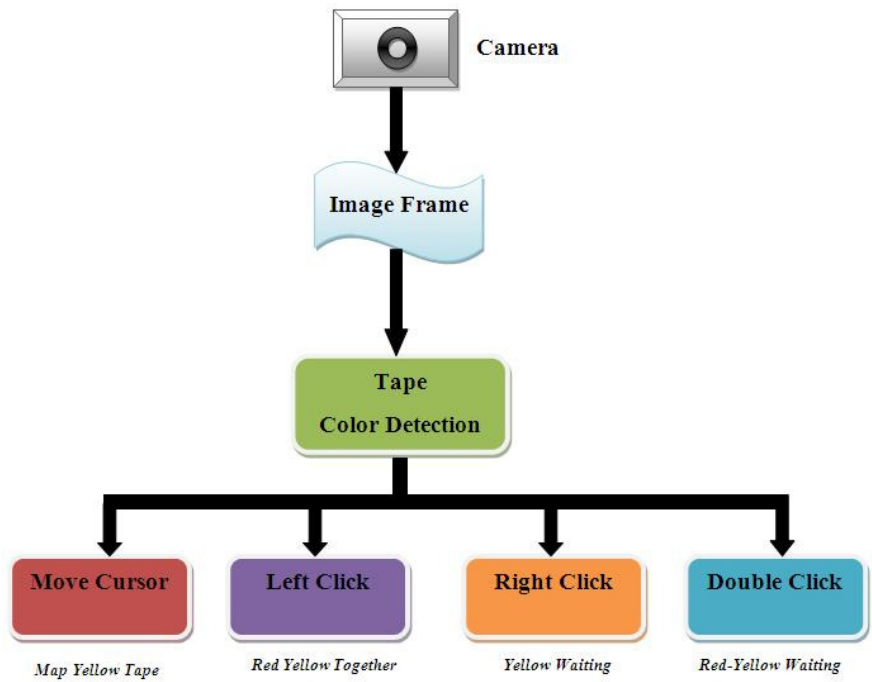

Figure 1: System Architecture

### 3.1. Hand Recognition and Colour Tape Detection

The first step of our system is to separate the potential hand pixels from the non-hand pixels. This can be done by background subtraction scheme which segments any potential foreground hand information from the non-changing background scene. At the system startup, a pair of background images is captured to represent the static workspace from camera view. Subsequent frames then use the appropriate background image to segment out moving foreground data. [4]

After background subtraction, the process of skin segmentation is done. Here, a histogram-based skin classifier assigns each of the RGB pixels in the training set to either a 3D skin histogram or non-skin histogram. Given these histograms, the probability is computed that a given RGB color belongs to the skin or non-skin classes. [4]

The skin segmentation process outputs an image which is ready for detection of color tapes in the finger. For this an algorithm based on HSV color space is used which is very effective to select a certain color out of an image. The idea is to convert the RGB pixels into the HSV color plane, so that it is less affected to variations in shades of similar color. Then, a tolerance mask is used over the converted image in the saturation and hue plane. The resulting binary image is then run through a convolution phase to reduce the noise introduced. [10]

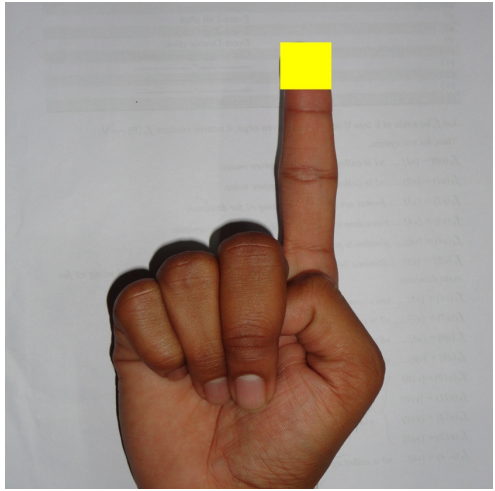

Figure 2: Yellow colour tape for cursor movement

## 3.2 Mouse Cursor Movement

We are using the index finger with yellow colour tape as a cursor controller to control mouse cursor movement. Two different approaches for moving the mouse cursor can be used. The first method is position mapping the index finger on a camera screen to a desktop screen position. But this method incurs a problem. If the resolution of the desktop window is greater than the camera resolution, then the cursor position cannot be accurate because while converting camera resolution to the desktop window resolution, intermediate values are lost. The expected ratio of jumping pixel is up to 4 pixels. The second method is known as weighted speed cursor control. Here the difference of the finger of the current image and the previous image is found and the distance between the two is computed. Next, the mouse cursor is moves fast if the gap between the two finger images (current and previous frame) is far or, if the gap is close then the cursor moves slow. There is a problem associated with this algorithm also. Some machines which cannot achieve image processing more than 15 fps do not work smoothly because computing the image center and the hand shape takes time. In this paper, we are concerning the first method which uses absolute position of finger tips because it is more accurate than the second method. [1]

## 3.3. Click Events

The click events for the mouse are mapped with different hand gestures. The idea focuses on processing the distance between the two coloured tapes in the fingers. The click events are detailed in the subsequent sub points.

### 3.3.1 Left Click

At the very first step, the system records the distance (say D) between the yellow and red tapes in the index finger and the thumb respectively. Here, the index and thumb must be apart as much as possible so as to get maximum distance (Figure: 3). This distance is regarded as the threshold distance for the event. Now, as the thumb moves towards the index finger, the distance between the finger tips or in other words, the distance between yellow and red tapes is decreases. In the second step, when the thumb is close to the index finger the system records the reduced distance (say D') between them (Figure: 4). When the distance between the tapes is reduced to D' or less we consider the event as the left click event of the mouse cursor.

Thus mathematically,

**D' < D**

Suppose the distance between the tapes at any time is **d** then for **left click** event

**d ≤ D'**

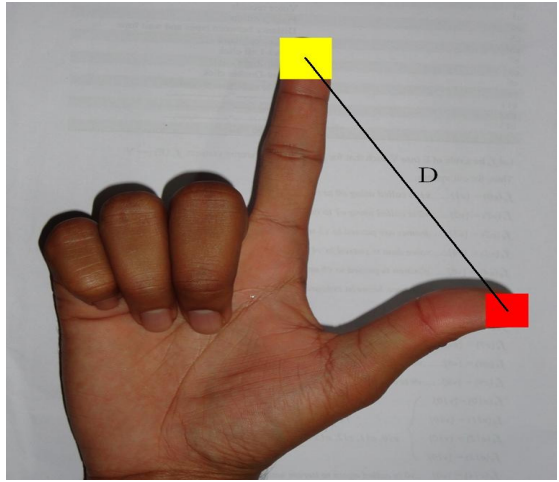

Figure 3: Initialisation of threshold distance(D)

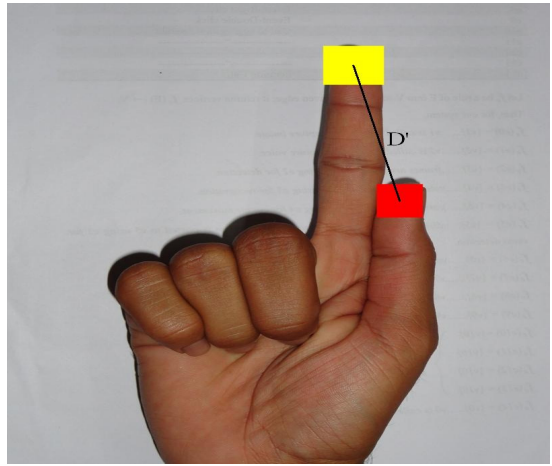

Figure 4: Reduced distance (D')

### 3.3.2 Right Click

The right click event of the cursor is simulated using the concept of waiting time. If the yellow tape on the index finger is waiting for 7 seconds(say) in front of the camera pointing at the same location, then the event is recognised as the right click event of the mouse cursor. Here, the distance between the red and yellow tapes should be between D and D' respectively. The required hand gesture is depicted in Figure: 3.

Thus, for **right click** event

**D' < d ≤ D**

**Waiting time = 7 sec.**

### 3.3.3 Double Click

The double click event of the cursor is also simulated in the same way as the right click event considering the waiting time. The only difference is that the finger gesture used for double click is as shown in Figure: 4. If both the colour tapes are waiting for the time 7 seconds (say) and the distance between the colour tapes is D' (reduced distance) or less then the event is recognised as double click event of the mouse cursor.

Thus, for **double click** event

$$d \leq D'$$

**Waiting time = 7 sec.**

## 4. DISCUSSIONS

In this system, we have proposed to use colour tapes on the fingers and all other functions can be done considering the relative distance of the tapes and the waiting time. This method has a greater efficiency over all other methods used earlier in this regard, where bare finger tips are used. Finger tip detection algorithms are not much effective as the colour of the tip of the finger cannot be differentiated from the colour of hand. This requires use of complex algorithms. To avoid such complex algorithms and make our system quick enough for real time computation, we have proposed use of colour tapes on the finger tips. This completely distinguishes the finger tip from the rest part of the hand. This distinction makes the colour detection algorithm [3] detect the tip quite easily and map it for cursor movement. Thus, the complexity of computation is reduced and overall results are improved.

## 5. APPLICATIONS

This technology can be used in robotics, gaming and developing systems which could understand human behaviour based on their way of interaction.

## 6. CONCLUSIONS

The system that we have proposed will completely revolutionize the way people would use the computer system. Presently, the webcam, microphone and mouse are an integral part of the computer system. Our product which uses only webcam would completely eliminate the mouse. Also this would lead to a new era of Human Computer Interaction (HCI) where no physical contact with the device is required.

## ACKNOWLEDGEMENTS

We would like to thank Prof. N. R. Talhar our guide in helping us to develop this concept. We would also like to thank Prof. S. V. Athawale for guiding us how to write the research papers. We would also thank the researchers, working in this field who in one way or another guided us in achieving our goals. We would also like to express our appreciation and gratitude to all other researchers at A.I.S.S.M.S College of Engineering, Pune who were kind enough to share their views with us and offered some suggestions in improving this idea.

## Authors

**Kamran Niyazi**  is currently pursuing B.E in Computer  Engineering at AISSMS College of Engineering under University of Pune, India. His research interest areas are Computer Networks, Image Processing, Database Management Systems and Data Structures.

**Vikram Kumar** is currently pursuing B.E in Computer  Engineering at AISSMS College of Engineering under University of Pune, India. His research interest areas are Image Processing and Algorithm Analysis.

**Swapnil Mahe** is currently pursuing B.E in Computer  Engineering at AISSMS College of Engineering under University of Pune, India. His research interest areas are Image Processing, Software Architecture and Design and Analysis of Algorithm.

**Swapnil Vyawahare** is currently pursuing B.E in Computer   Engineering at AISSMS College of Engineering under University of Pune, India. His research interest areas are Image Processing, Computer Neetworks and Design and Analysis of Algorithm.